\documentclass[letterpaper, 10 pt, conference]{ieeeconf}

\IEEEoverridecommandlockouts
\usepackage{graphics}
\usepackage{epsfig}
\usepackage{mathptmx}
\usepackage{times}
\usepackage{amsmath}
\usepackage{amssymb}
\usepackage{booktabs}
\usepackage[table]{xcolor}
\definecolor{darkred}{rgb}{0.55, 0.0, 0.0}
\definecolor{darkgreen}{rgb}{0.0, 0.5, 0.0}
\usepackage{stmaryrd}
\usepackage{algorithm}
\usepackage{algpseudocode}
\usepackage{wasysym}
\usepackage{multirow}
\usepackage{gensymb}
\title{\LARGE \bf
3D Can Be Explored In 2D : Pseudo-Label Generation for LiDAR Point Clouds Using Sensor-Intensity-Based 2D Semantic Segmentation
}

\author{Andrew Caunes$^{1}$$^{2}$, Thierry Chateau$^{1}$, Vincent Frémont$^{2}$
\thanks{$^{1}$Logiroad}
\thanks{$^{2}$LS2N - Ecole Centrale de Nantes}
}

\begin{document}

\maketitle
\thispagestyle{empty}
\pagestyle{empty}

\begin{abstract}
Semantic segmentation of 3D LiDAR point clouds, essential for autonomous driving and infrastructure
management, is best achieved by supervised learning, which demands extensive annotated datasets and faces the 
problem of domain shifts. We introduce a new 3D semantic segmentation pipeline that leverages aligned scenes and 
state-of-the-art 2D segmentation methods, avoiding the need for direct 3D annotation or reliance on additional 
modalities such as camera images at inference time. Our approach generates 2D views from LiDAR scans colored
by sensor intensity and applies 2D semantic segmentation to these views using 
a camera-domain pretrained model. The segmented 2D outputs are then back-projected onto the 
3D points, with a simple voting-based estimator that merges 
the labels associated to each 3D point. Our main contribution is a global 
pipeline for 3D semantic segmentation requiring no prior 3D annotation 
and not other modality for inference, which can be used for pseudo-label generation. 
We conduct a thorough ablation study and demonstrate the potential 
of the generated pseudo-labels for the Unsupervised Domain Adaptation task.
\end{abstract}

\section{INTRODUCTION}

\label{introduction}

Semantic segmentation of 3D point clouds, gathered by LiDAR sensors, is pivotal in various domains, 
notably in Autonomous Driving (AD) and infrastructure management. This segmentation is crucial for tasks such as environment 
understanding and object interaction—key components for autonomous 
system navigation and decision-making. However, the majority of 3D 
semantic segmentation work, particularly for LiDAR, has been tailored to AD.
This specialization focuses on dynamic objects like vehicles, often neglecting the full 
potential of dense scenes of aligned point clouds for static objects. Moreover, the emphasis 
on real-time processing limits the scope of computational resources, 
impacting the reach of segmentation achievable.

The highest performance in 3D semantic segmentation has been achieved through supervised learning \cite{wu_point_2023} \cite{choy_4d_2019}, 
requiring extensive annotated data. 
This approach is notably hindered by the significant resources required for annotation, as well as the inherent difficulty 
in annotating complex 3D data \cite{caesar_nuscenes_2020} \cite{behley_semantickitti_2019}. Additionally, 
existing methods often suffer from substantial \textit{domain shift} when applied 
to diverse or unseen environments, which hampers the model's ability to generalize across diverse environments.

In response to these challenges, various approaches have been proposed to avoid or reduce the need for point cloud annotation.
Weak and semi-supervised learning strategies aim to maximize performance with minimal annotated data \cite{lee_pseudo-label_2013}. 
Domain adaptation methods aim to overcome the domain shift problem, enabling the use of existing datasets \cite{corral-soto_domain_2023} \cite{rochan_unsupervised_2022} \cite{michele_saluda_2023}. 
Active learning strategies aim to minimize this need by carefully selecting data samples for annotation. 
While all of these methods have shown promising results, they still require a certain level of annotation, either for the target domain or for the source domain.
Encouraged by the rapid advancements in 2D image segmentation in recent years, novel solutions
 have been proposed to transfer knowledge
from 2D to 3D, thus circumventing the need for 3D annotations as well as avoiding the 3D domain shift problem 
\cite{sautier_image--lidar_2022} \cite{liu_segment_2023} \cite{yang_sam3d_2023} \cite{huang_segment3d_2023}. 
Our approach, while falling into this category, uniquely does not rely on additional modalities such as camera images for inference.

We propose a pseudo-label generation pipeline leveraging aligned scenes, higher computing time and state of the art 2D methods
to generate high quality labels. These labels can then be used either on their own or as training material e.g. in domain adaptation.
The only requirements are that input point clouds must be provided in sequences and either a 
pretrained 2D image segmentation model or a 2D annotated dataset must be available for the desired classes at training time. 
It should be highlighted that our method is optimal for use
with static classes, unaffected by movement artifacts in aligned scenes. We plan to investigate
solutions for dynamic classes in future work.\\
First, all scans of a 3D scene are aligned using input sensor poses which are either provided or computed. 
A large number of views of the scene are then generated as 2D RGB images, using only the LiDAR intensities as colors. This differs from
prior arts where camera images are used for direct inference or 
point coloring \cite{mascaro_diffuser_2021} \cite{wang_ldls_2019} \cite{genova_learning_2021}. 
Ideally, the views of the
point cloud as well as the augmented training images of the 2D model are selected with the objective of 
minimizing the domain shift between them.
These views are then processed through the 2D segmentation model, and the resulting labels are projected 
onto the 3D points. The 3D segmentation
masks are accumulated, effectively being used as \textit{``votes''}. Each point's final class is then determined 
based on a simple election estimator.

In this paper, we make two key contributions to the field of 3D semantic segmentation. First, we show that direct inference of 2D image models
on views of raw dense point clouds with only sensor-intensity-based coloring can be used for 3D semantic segmentation
without requiring prior 3D annotations or other modality at inference time. 
Second, we show that this method can be used to produce quality pseudo-labels. 
We demonstrate the potential of our pseudo-labels firstly by comparing them to ground-truth labels and secondly by
using them in the 3D Unsupervised Domain Adaptation (UDA) setting and comparing the resulting models to state-of-the-art methods.

\section{RELATED WORKS}
\label{related_works}

Most of the existing approaches for 3D semantic segmentation utilize 3D annotations, either in the source or the target domain. 
When 3D labels are available in the target domain, fully-supervised 3D models achieve the best performance 
\cite{choy_4d_2019} \cite{wu_point_2023} with models either 
using sparse 3D convolutions or 2D Bird's Eye View representations.

When the target data comes with few annotated samples, semi-supervised methods can be used to leverage unlabelled target data. In that case, it has been shown that self-training on 
pseudo-labels can significantly improve performance by approximating entropy maximization 
\cite{yang_st3d_2021} \cite{lee_pseudo-label_2013}. 

Some domain adaptation methods also leverage a few target labels to improve performance \cite{corral-soto_domain_2023}.

With only source domain 3D labels, unsupervised domain adaptation methods have been proposed to mitigate the domain shift problem \cite{michele_saluda_2023} 
\cite{rochan_unsupervised_2022}, \cite{vu_advent_2019}, \cite{gebrehiwot_t-uda_2023}, \cite{sun_deep_2016}, \cite{lee_sliced_2019}.

Most recently, zero-shot learning methods have been shown to perform well in 3D semantic segmentation. 
\cite{wang_transferring_2023} and \cite{lu_see_2023} align 2D, 3D and language features in a common latent space in order to segment
unseen classes in a target domain.

Substantial progress has been made recently towards 3D semantic segmentation without any 3D annotation. This implies using other 
modalities for training and / or inference. Most recently, self-supervised methods have been proposed where a model is pretrained
on unlabelled data using a pretext task to obtain relevant feature representations. 
2D-to-3D knowledge distillation \cite{sautier_image--lidar_2022} 
\cite{liu_segment_2023} leverages 2D image models to directly teach a student 3D model. Though very efficient, this approach still requires 
downstream fine-tuning on 3D labelled data to achieve a given task.

We propose to segment multiple 2D views of a 3D point cloud, therefore requiring
no 3D annotation but only 2D annotations at training time. Close to our proposed method,
 \cite{seeland_multi-view_2021} provides a comprehensive review of 
 the state-of-the-art techniques that use multiple views of a
scene for inference with Convolutional Neural Networks (CNN).
These methods can be divided into different categories, 
depending on the stage at which information from multiple views is fused.
Our method falls into the category that uses \textit{score fusion}, which involves 
performing inference independently on each view first and 
then fusing the resulting scores. 
The choice of the fusion method is investigated in \cite{seeland_multi-view_2021}
where they experiment with adding, compounding and taking a maximum of the scores.
The summing estimator can be seen as a simple majority election system where each 
projected classification of a point (or pixel) constitutes a vote.
\cite{pellis_2d_2022} uses this method with colored point clouds for heritage building 3D segmentation. 
The method proposed in \cite{genova_learning_2021} is very close to ours except that
posed camera RGB images are used for inference instead of generated views of the point cloud.
\cite{wang_ldls_2019} and \cite{mascaro_diffuser_2021} use a more complex 
graph-based diffusion mechanism accounting for neighboring points
to project and fuse the 2D labels to 3D. 
\cite{robert_learning_2022} and \cite{peters_semantic_2023} experiment with 
using neural networks to correspond 2D features to 3D,
a process that requires 3D annotations for training.
Very recently, \cite{yang_sam3d_2023} and \cite{huang_segment3d_2023} applied
 a 2D-to-3D label propagation method to non-semantic 
segmentation labels from Vision Foundation Models (VFM) such as
 Segment-Anything \cite{kirillov_segment_2023}.
To the best of our knowledge, our method is the first to perform 2D-to-3D 
label propagation to generated 2D views thereby eliminating the need for camera images
at inference time.
We experiment with various fusion estimators and show that a 
soft summing estimator can yield satisfactory results.

\section{METHOD}
\label{method}

\begin{figure*}[thpb]
      \centering
      \includegraphics[width=0.96\textwidth]{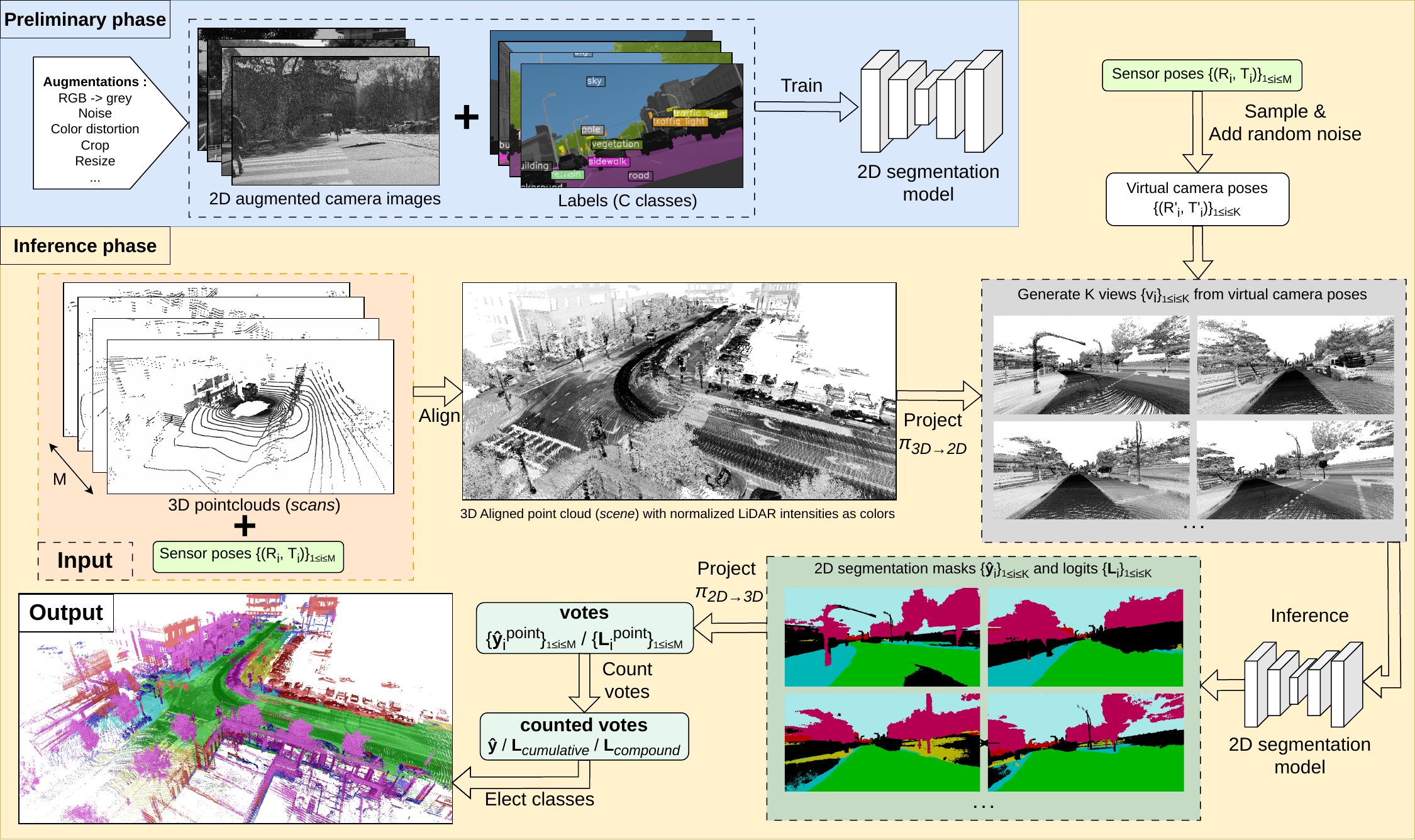}
      \caption{Pipeline. As a preliminary step, a 2D semantic segmentation model is trained on augmented camera images.
      An input sequence of LiDAR 3D scans and sensor poses is aligned 
      and colorized using sensor intensity. A large number of 2D greyscale images are then 
      generated from virtual camera poses along the sensor's trajectory. 
      The 2D model is applied to obtain segmentation logits and labels for each view 
      which are then projected back onto the 3D points. These \textit{votes} are
      counted and each point is assigned a class based on the chosen election estimator.}
      \label{img:method}
\end{figure*}

Our proposed pipeline is illustrated in Fig. \ref{img:method}. As a preliminary step
, a trained 2D image segmentation model must be obtained. Ideally, the model is 
trained using image augmentations in order to minimize the domain gap 
with the target images that will be generated subsequently.
We train our own 2D models in Section \ref{sec:experiments} and include 
the influence of the augmentations in the ablation study.

Given a sequence of raw LiDAR scans $S=\{s_1,s_2,\ldots,s_M\}$, where 
$\forall m,\; s_m \in \mathbb{R}^{N_m \times 4}$ (each point has 3 coordinates 
and an intensity), we colorize each point \(p_{j,m}\) in scan \(s_m\) based on its intensity 
(LiDAR reflectivity). We apply two transformations to these intensities with the objective to 
increase contrast and to close the domain gap between the training images of the 2D model and the
generated views. The intensity \(I_{j,m}\) of each point is first clipped to a range
 \([\beta_{min}, \beta_{max}]\) to mitigate the effect of extremely high or low values:
\[
I'_{j,m} = \min(\max(I_{j,m}, \beta_{min}), \beta_{max}),
\]
The clipped intensity is then scaled to a range \([\eta_{\text{min}},\eta_{\text{max}}]\).
\[
I''_{j,m} = \frac{I'_{j,m} - \min_j(I'_{j,m})}{\max_j(I'_{j,m}) - \min_j(I'_{j,m})} 
\times (\eta_{\text{max}} - \eta_{\text{min}}) + \eta_{\text{min}}.
\]
Each scan \(s_m\) is then aligned into a
common reference frame using transformation \(T_m\), obtained from sensor poses.
\[
S_{dense} = \bigcup_{m=1}^{M} T_m(s_m).
\]
When sensor poses are not readily available, they can be estimated using odometry methods such as 
Iterative Closest Point (ICP). Experimentation on our own data shows that a 64-beam LiDAR 
sensor with a 20Hz acquisition rate can be aligned automatically using \cite{vizzo_kiss-icp_2023} 
with enough precision for our method to work with satisfying performance. 

From the sensor poses, $K$ multiple virtual camera poses \(\{(R'_i, T'_i)\}_{1\le i \le K}\) are 
generated by simulating a vehicle-like perspective at different positions 
and orientations around \(S_{dense}\), as described in Alg. \ref{alg:generate_views}. 
\begin{algorithm}
    \caption{Generate virtual camera poses}
    \label{alg:generate_views}
    \begin{algorithmic}[1]
    \ForAll{$i$ in $\llbracket 1, K \rrbracket$}
        \State $R'_i$, $T'_i$ = $RandomChoice(\{R_i,T_i\}_{1<i<\mathbf{M}})$
        \State $R'_i$ = $R'_i \cdot Rotation_Y(\mathcal{U}(-\theta, \theta))$
        \State $T'_i$ = $T'_i$ + $\mathcal{U}(-\lambda, \lambda) * (R'_i)_X$
        \State $T'_i$ = $T'_i$ + $\mathcal{U}(-\lambda, \lambda) * (R'_i)_Z$
        \State $T'_i$ = $T'_i$ + $\mathcal{U}(-\gamma, \gamma) * (R'_i)_Y$
    \EndFor
\end{algorithmic}
\end{algorithm} 
We add random noise to the original poses to enhance the diversity of viewpoints.
We then obtain views $V = \{v_1, v_2, \ldots, v_K\}$
corresponding to 2D projections of the 3D scene:
\[
\forall k, \; v_k = \pi_{3D\rightarrow 2D}(S_{dense};\; R'_k, T'_k),
\]
During the inference stage, for each generated view \(v_k\), the 2D segmentation model
\(f_{2D}\) outputs logits $\mathbf{L}^{\text{pixel}}_k$ 
and a segmentation mask $\mathbf{\hat{y}}^{\text{pixel}}_k$.
\begin{align*}
\mathbf{L}^{\text{pixel}}_k &= f_{2D}(v_k) \in \mathbb{R}^{width \times height \times C}, \\
\mathbf{\hat{y}}^{\text{pixel}}_k &= \arg\max(\text{Softmax}(f_{2D}(v_k))) \in \{0,1\}^{width \times height \times C}
\end{align*}
The projection step maps the predicted 2D segmentation masks $\mathbf{\hat{y}}^{\text{pixel}}_k$ and 
logits $\mathbf{L}^{\text{pixel}}_k$ back onto the 3D points in $S_{dense}$. 
For each view \(v_k\), we obtain point logits $\mathbf{L}^{\text{point}}_k$
and a point segmentation mask $\mathbf{\hat{y}}^{\text{point}}_k$
by projecting using the pose $(R'_k, T'_k)$ :
\begin{align*}
    \mathbf{L}^{\text{point}}_k &= \pi_{2D \rightarrow 3D}(\mathbf{L}^{\text{pixel}}_k;\; R'_k, T'_k) \in \mathbb{R}^{M \times C}, \\
    \mathbf{\hat{y}}^{\text{point}}_k &= \pi_{2D \rightarrow 3D}(\mathbf{\hat{y}}^{\text{pixel}}_k;\; R'_k, T'_k) \in \{0,1\}^{M \times C}
\end{align*}
    
We accumulate and compound the obtained point segmentation masks and logits 
from all views, a process that can be understood as counting \textit{votes}. We
experiment with three different fusion methods (estimators):
\begin{align*}
\mathbf{\hat{y}}^{\text{point}} &= \sum_{k=1}^{K} \mathbf{\hat{y}}^{\text{point}}_k \in \mathbb{N}^{M \times C},, \\
\mathbf{L}^{\text{point}}_{\text{cumulative}} &= \sum_{k=1}^{K} \mathbf{L}^{\text{point}}_k \in \mathbb{R}^{M \times C}, \\
\mathbf{L}^{\text{point}}_{\text{compound}} &= \prod_{k=1}^{K} \mathbf{L}^{\text{point}}_k \in \mathbb{R}^{M \times C}.
\end{align*}
A final class is then picked for each point according to the chosen election estimator:
\begin{itemize}
    \item \textbf{Summing hard votes} :
\end{itemize}
\begin{align*}
    \hat{y}^{\text{point}} = \arg\max(\mathbf{\hat{y}}^{\text{point}}) \in \llbracket 1,C \rrbracket^{M}
\end{align*}
\begin{itemize}
    \item \textbf{Summing Soft Votes} :
\end{itemize}
\begin{align*}
    \hat{y}^{\text{point}} = \arg\max(\mathbf{L}^{\text{point}}_{\text{cumulative}}) \in \llbracket 1,C \rrbracket^{M}
\end{align*}
\begin{itemize}
    \item \textbf{Compounding Soft Votes} :
\end{itemize}
\begin{align*}
    \hat{y}^{\text{point}} = \arg\max(\mathbf{L}^{\text{point}}_{\text{compound}}) \in \llbracket 1,C \rrbracket^{M}
\end{align*}

\section{EXPERIMENTS}
\label{sec:experiments}

To demonstrate the effectiveness of our approach, we generate pseudo-labels 
for the nuScenes autonomous driving dataset \cite{caesar_nuscenes_2020}, and then evaluate the quality of
these labels in two ways. First, we compare the pseudo-labels generated directly to the ground-truth annotations.
We conduct an ablation study to verify the importance of various features of our pipeline.
Second, we utilize the obtained pseudo-labels to train 3D semantic segmentation models in the 
Unsupervised Domain Adaptation setting and compare their performance to that of 
state-of-the-art methods.

\subsection{Pseudo-labels Generation}
\subsubsection{Implementation details}
For our first experiment, we generate pseudo-labels on the nuScenes autonomous 
driving dataset \cite{caesar_nuscenes_2020} and compare them
to the ground-truth annotations. nuScenes is a large-scale dataset
providing semantic segmentation annotations for more than 34,000 LiDAR scans acquired
using a 32-beam LiDAR sensor at 20Hz. The dataset is split into 1,000 scenes, with
850 annotated scenes of which 150 are for validation. There are a total of 32 
annotated classes for 3D semantic segmentation, but we will focus on a subset of 5
classes: \textit{driveable surface}, \textit{sidewalk}, \textit{manmade}, 
\textit{vegetation} and \textit{terrain}. We select these classes for their 
static nature which make them unaffected by motion-related artifacts in an 
aligned sequence of scans. 
For each scene, nuScenes provides 
the full sequence of LiDAR scans of which about 1/10 is annotated.
The sensor poses which we use to align the scans are also provided.

We use the Intersection over Union (IoU) metric for the comparison. 
We compute the IoU for each class as well as the mean
IoU (mIoU) over all classes. We run our pipeline over all 850 annotated
scenes of the nuScenes dataset.
Points further than 30 meters on each side and 10 meters in height are 
cropped for the comparison since the generated views do not explore 
these areas. In the future, we plan to adress this issue by
generating more diverse poses and adapting the 2D model
to other domains than the classical vehicle perspective. 
The \textit{sidewalk} and \textit{terrain}
classes are also merged for this comparison, 
as they are very close in LiDAR intensity domain. 

We sample $K=600$ poses from the original sensor poses for 
each scene and add random noise to them following 
Alg. \ref{alg:generate_views} with 
$\theta=30\degree$, $\lambda=1m$, $\gamma=1m$.
Examples of the obtained view can
be seen in Fig. \ref{img:method}. 
We set a limit on the maximum depth for 
the back-projection to 30 meters and a minimum depth of 1 meter.
We discuss the influence of all parameters in Table \ref{tab:ablation}. 

We train our own 2D semantic segmentation model. For this, we require a dataset
annotating RGB images with the same classes as our target dataset. We use
the Mapillary Vistas dataset \cite{neuhold_mapillary_2017} which provides 20,000
annotated street images of high diversity. We believe this diversity can help bridge the domain gap 
between camera images and generated views. For the same reason, we augment
the images in multiple ways including random cropping, random scaling, 
random horizontal flipping, random color jittering, random gaussian noise,
random gaussian blur and rotations. Most importantly, we convert the RGB images
to greyscale images to match the LiDAR intensity domain. We also experiment with
the influence of these augmentations in Table \ref{tab:ablation}. Examples of the images
can be seen in Fig. \ref{img:method} \\
We train a Mask2Former \cite{cheng_masked-attention_2022} architecture
with a swin-l backbone \cite{liu_swin_2021} using \cite{mmsegmentation_contributors_openmmlab_2020} for 9 epochs 
with a batch size of 1 on 4 Nvidia V100 GPUs. We use a learning rate of 0.0001
with polynomial decay and a momentum of 0.9 and a weight decay of 0.0001.
\subsubsection{Results}

\begin{figure}[thpb]
      \centering
      \includegraphics[width=0.5\textwidth]{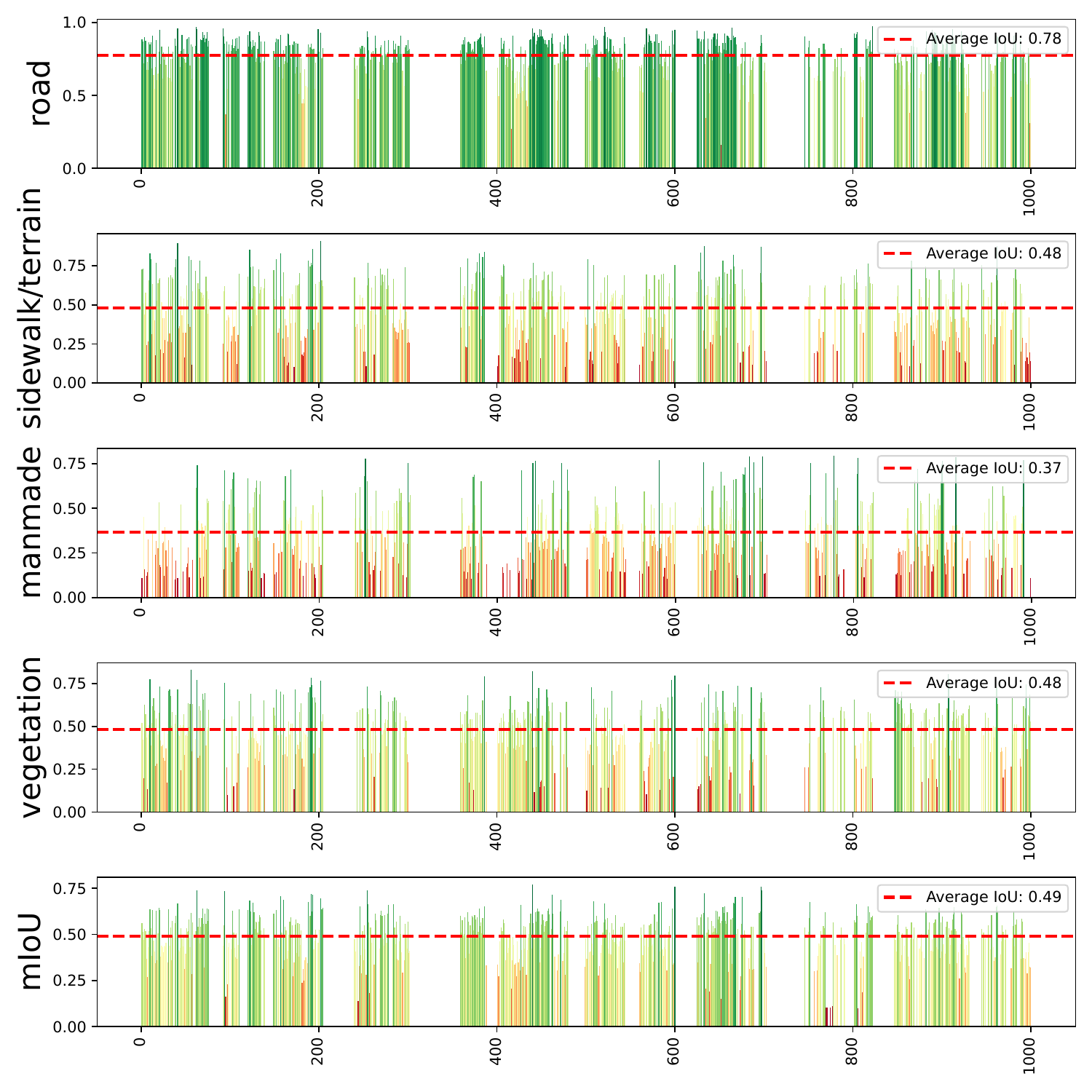}
      \caption{Comparison of the pseudo-labels generated on the nuScenes dataset \cite{caesar_nuscenes_2020}.
      IoU per class and mIoU for all annotated scenes. Transformations such as cropping are applied to 
      obtain a relevant comparison.}
      \label{img:pseudo_labels_bar_plots}
\end{figure}

The comparison of our pseudo-labels with ground-truth annotations is displayed in 
Fig. \ref{img:pseudo_labels_bar_plots}.
The best performance is obtained for the \textit{driveable surface} class with an average IoU of 0.81.
This is explained by the higher contrast between the road and the rest of the scene.
The \textit{manmade} class is the least well predicted with a mIoU score of 0.37. This and other
low scores may be caused by three factors: the occlusion and depth problem and the lack of exploration. 
Indeed, buildings tend to be further away from the sensor. This leads to some points being 
ouf of reach for many views given our depth limit, and exploration isn't sufficient because 
only views close to the sensor trajectory are rendered. There is also an 
occlusion problem where background hidden points receive votes from 2D segmentation 
masks of foreground objects. 
This problem is most prevalent for buildings which are further away. 
We plan to address these issues in the future by using depth maps to deal with occlusions and
by generating more diverse poses.
Note that some scenes yield notably low mIoU scores because the ego vehicle remains stationary. 
This results in an unexplored environment and therefore a sparse aligned point cloud.

We conduct an ablation study to verify the influence of various features of our pipeline in Table \ref{tab:ablation}. 
Each line corresponds to an experiment, the first experiment (a) serving as a reference. 
Related experiments from (b) to (h) are grouped by color and compared to (a).
Various settings are explored where the presence of a feature is indicated by a checkmark. 
Presented features are, respectively, \textit{number of generated views} (with varying $K$), \textit{2D model image augmentation}
(either all augmentations or none), \textit{randomization of the virtual camera poses} (either
low or high values for $\theta$, $\lambda$ and $\gamma$), 
\textit{election estimator}.
The green area shows that the number of generated views has a significant impact on performance, 
with the mIoU score dropping more than 86\% when decreasing the number of views from 600 to 100. 
Removing the image augmentations for the 2D model has a modest negative impact. 
High variation in the poses generation and the choice of election estimator, 
seen in the red and yellow areas, seem to have a strong 
impact on performance. In particular, ``summing soft votes'' seems to be
the best election estimator. 
Interestingly, \cite{seeland_multi-view_2021} found that the ``coumpounding soft votes''
election estimator yielded the best results, while we observe a large drop in performance. This may be caused
by a missing prior normalization of the scores.

\begin{table}[ht]
\centering
\caption{Ablation study.}
\label{tab:ablation}
\begin{tabular}{p{0.2cm}|p{0.3cm}p{0.cm}p{0.cm}p{0.cm}p{0.2cm}p{0.2cm}p{0.2cm}p{0.cm}p{0.cm}p{0.cm}cc}
    \bottomrule
    &\tiny \multirow{2}{*}{\textbf{K=100}} &\tiny \multirow{2}{*}{\textbf{300}} &\tiny \multirow{2}{*}{\textbf{600}} &\tiny \multirow{2}{*}{\textbf{900}} & \tiny \textbf{img. aug}. &\tiny \textbf{rdm. vws.} &\tiny \textbf{high rdm. vws.} &\tiny \textbf{$+$ hrd. vts.} & \tiny \textbf{$+$ sft. vts.} &\tiny \textbf{$\times$ sft. vts.} &\tiny \multirow{2}{*}{\textbf{mIoU}} & \tiny \multirow{2}{*}{\textbf{$\nearrow\%$}}\\
    \hline
    \rowcolor{gray!20} (a)  &            &            &            & \checkmark & \checkmark & \checkmark &            & \checkmark &            &            & 50.06    & \textit{ref}       \\
    \hline
    \rowcolor{green!5}  (b) & \checkmark &            &            &            & \checkmark & \checkmark &            & \checkmark &            &            & 6.89     & \textcolor{darkred}{-86.23}     \\
    \rowcolor{green!5}  (c) &            & \checkmark &            &            & \checkmark & \checkmark &            & \checkmark &            &            & 39.5     & \textcolor{darkred}{-21.09}       \\
    \rowcolor{green!5}  (d) &            &            & \checkmark &            & \checkmark & \checkmark &            & \checkmark &            &            & 49.95    & \textcolor{darkred}{-0.22}      \\
    \hline
    \rowcolor{blue!5} (e)   &            &            &            & \checkmark &            & \checkmark &            & \checkmark &            &            & 49.77    & \textcolor{darkred}{-0.58}      \\
    \hline
    \rowcolor{red!5}  (f)   &            &            &            & \checkmark & \checkmark &            & \checkmark & \checkmark &            &            & \textbf{53.14}& \textcolor{darkgreen}{+6.15} \\
    \hline
    \rowcolor{yellow!5} (g) &            &            &            & \checkmark & \checkmark & \checkmark &            &            & \checkmark &            & 53.12     & \textcolor{darkgreen}{+6.12}     \\
    \rowcolor{yellow!5} (h) &            &            &            & \checkmark & \checkmark & \checkmark &            &            &            & \checkmark & 35.77 & \textcolor{darkred}{-28.54} \\
    \toprule
\end{tabular}
\end{table}
    
\subsection{Training 3D Models for Domain Adaptation}
\subsubsection{Implementation Details}
As a more practical approach for evaluating the quality of our method and pseudo-labels, 
we conduct experiments in the Unsupervised Domain Adaptation setting. 
A 3D model is pretrained on a labelled source semantic segmentation dataset and 
then re-trained or fine-tuned on an unlabelled target dataset using the pseudo-labels generated 
by our method. We use the SemanticKITTI dataset \cite{behley_semantickitti_2019} as the source dataset 
as it also provides 3D semantic segmentation for the classes of interest.
We compare our method to two approaches: weak-supervision baselines, which are models pretrained 
on SemanticKITTI and fine-tuned directly on a subset of nuScenes' annotations, and state-of-the-art
UDA methods. 

For the unsupervised domain adaptation methods, we compare our approach to SWD \cite{lee_sliced_2019}, (M+A)Ent \cite{vu_advent_2019},
CORAL \cite{sun_deep_2016}, UDASSGA \cite{rochan_unsupervised_2022} and the recent T-UDA \cite{gebrehiwot_t-uda_2023}. For a fair comparison, 
we follow \cite{gebrehiwot_t-uda_2023} and use a Minkowski U-net architecture \cite{choy_4d_2019}. The results for 
the compared methods are taken from \cite{gebrehiwot_t-uda_2023}.
Note that the \textit{Naïve} setting (''no da'') in \cite{gebrehiwot_t-uda_2023} obtains a mIoU score of 34.18 on the classes of interest which is
significantly higher than our own \textit{Naïve} implementation. We believe this may be due to a difference 
in the augmentations used during pretraining on SemanticKITTI, but the comparison should still hold since
our results are lower.
We use \cite{mmdetection3d_contributors_openmmlabs_2020} and train each model 
for 36 epochs on the source dataset, then fine-tune 
on the target dataset for 36 epochs. We use a batch size of 8 for 
SemanticKITTI and 32 for nuScenes on a single Nvidia V100 GPU. 

\subsubsection{Results}

The results are displayed in Table \ref{tab:da}. The weak-supevision baselines and UDA methods 
are at the top and middle of the table respectively. We split our results in two experiments depending
on whether the model is pretrained on SemanticKITTI or not. The best overall results are in bold while the 
best and second best unsupervised results are in blue and red respectively.
As a first observation, both results are close 
to each other. This is satisfactory as it shows that our method can be used without the need for
3D semantic segmentation annotations. Secondly, the pretrained method is only 6.8 mIoU points below 
the 1\% weak-supervision baseline, which, for nuScenes, represents about 300 manually annotated scans.
Interestingly, in both settings, our method performs the worst on the \textit{vegetation} class, although the pseudo-labels
obtained better IoU than for the \textit{manmade} class on average in Fig. \ref{img:pseudo_labels_bar_plots}. 
We believe this is due to the fact that even with better annotations, 
identifying leafs in a sparse 3D scan is a harder task than simple plane detection
which can be used to segment a large proportion of the \textit{manmade} class. 

Compared to state-of-the-art methods in unsupervised domain adaptation, 
our approach is on-par with T-UDA \cite{gebrehiwot_t-uda_2023}
and reaches better mIoU than all other methods in the pretrained setting. Notably, this last statement 
is still true even without pretraining on SemanticKITTI.

\begin{table}[ht]
    \centering
    \caption{IoU per class and mIoU for the domain adaptation setting.}
    \label{tab:da}
    \begin{tabular}{lp{0.4cm}p{0.7cm}p{0.7cm}p{0.5cm}p{0.8cm}p{0.7cm}}
    \hline
     & road  & sidewalk & manmade & terrain & vegetation & mIoU \\
    \hline
    $SK \rightarrow NS$ & & & & & & \\
    \hline
    Naïve & 9.7 & 0.1 & 24.7 & 0.7 & 0.3 & 7 \\
    1\permil & 78.7 & 0.2 & 47.2 & 43.7 & 48.6  & 43.7 \\
    1\% & 90.2 & 44 & 79.9 & 70.1 & 56.8 & 68.2  \\
    10\% & \textbf{94.9} & \textbf{66.6} & \textbf{86.5} & \textbf{83} & \textbf{72} & \textbf{80.5}  \\
    \hline
    SWD \cite{lee_sliced_2019} & 80.7 & 26.5 & 60.2 & 30.1 & 43.9 & 48.28 \\
    (M+A)Ent \cite{vu_advent_2019} & 83.5 & 32.6 & 62.3 & 31.8 & 43.3 & 50.7 \\
    CORAL \cite{sun_deep_2016} & 82.6 & 27.1 & 56.7 & 27 & 55.3 & 49.74 \\
    UDASSGA \cite{rochan_unsupervised_2022} & 82.2 & 29.6 & 65.7 & 34 & \textcolor{red}{57.9} & 53.88 \\
    T-UDA \cite{gebrehiwot_t-uda_2023} & \textcolor{blue}{87.8} & \textcolor{blue}{45.8} & \textcolor{red}{72.6} & 46.1 & \textcolor{blue}{70.3} & \textcolor{blue}{64.52} \\
    \hline
    \rowcolor{blue!10} Ours & 86.2 & 30 & 67.9 & \textcolor{red}{64.9} & 22 & 54.23 \\
    \rowcolor{blue!10} Ours pt SK & \textcolor{red}{87.5} & \textcolor{red}{40} & \textcolor{blue}{73} & \textcolor{blue}{69} & 37.1 & \textcolor{red}{61.4} \\
    \hline
    \rowcolor{lightgray} full. sup. & 96.6 & 73.9 & 89.6 & 86 & 75 & 84.3 \\
    \end{tabular}
\end{table}

\subsection{Conclusion}

We have proposed a method for 3D semantic segmentation of sequences of point clouds requiring
zero 3D annotations. Unlike prior works, we do not require any additional modality such 
as camera images for inference but only 2D annotations for training. 
We demonstrated the potential of the method for producing pseudo-labels which can
be used in the Unsupervised Domain Adaptation setting to obtain competitive results.

While promising, our initial findings also highlight several challenges,
such as the occlusion problem and the limited domain in which 
the views can be generated without increasing the domain shift with the 2D dataset.
Future efforts will focus on mitigating these issues to enhance the robustness and applicability of our method.

Avenues for improvement also include extending our technique to other classes. In particular,
non-static classes such as vehicles and pedestrians could also be reached by adapting the 2D
model to the domain of generated views, which includes motion-related artifacts.

\section{acknowledgements}
This project was provided with AI computing and storage resources by GENCI at IDRIS thanks to the grant 2024- AD011012128 on the supercomputer Jean Zay's V100 partition.

\bibliography{bib}{}
\bibliographystyle{plain}

\end{document}